\pdfoutput=1

\documentclass[11pt]{article}

\usepackage{EMNLP2022}
\usepackage[font={normal}]{caption}

\usepackage{times}
\usepackage{latexsym}
\usepackage{wrapfig}

\usepackage[T1]{fontenc}
\usepackage[english]{babel}
\usepackage[utf8]{inputenc}

\usepackage{microtype}
\usepackage{listings}
\usepackage{inconsolata}
\usepackage{fancyvrb} 
\usepackage{multirow}
\usepackage{enumitem}
\usepackage{booktabs}
\usepackage{graphicx}
\usepackage{tabularx}
\usepackage{multicol}
\usepackage{expex}
\usepackage{xspace}
\usepackage[]{algorithm2e}
\usepackage{amsmath}

\definecolor{cb-black}      {RGB}{  0,   0,   0}
\definecolor{cb-blue-green} {RGB}{  0, 073, 073}
\definecolor{cb-green-sea}  {RGB}{  0, 146, 146}
\definecolor{cb-rose}       {RGB}{255, 109, 182}
\definecolor{cb-salmon-pink}{RGB}{255, 182, 119}
\definecolor{cb-purple}     {RGB}{ 73,   0, 146}
\definecolor{cb-blue}       {RGB}{ 0,  109, 219}
\definecolor{cb-lilac}      {RGB}{182, 109, 255}
\definecolor{cb-blue-sky}   {RGB}{109, 182, 255}
\definecolor{cb-blue-light} {RGB}{182, 219, 255}
\definecolor{cb-burgundy}   {RGB}{146,   0,   0}
\definecolor{cb-brown}      {RGB}{146,  73,   0}
\definecolor{cb-clay}       {RGB}{219, 209,   0}
\definecolor{cb-green-lime} {RGB}{ 36, 255,  36}
\definecolor{cb-yellow}     {RGB}{255, 255, 109}

\newcommand{\frechet}{$\delta_F$\xspace}
\newcommand{\pearson}{$r$\xspace}

\title{Universal and Independent: Multilingual Probing Framework for Exhaustive Model Interpretation and Evaluation}

\newcommand{\hse}{\heartsuit}
\newcommand{\mipt}{\spadesuit}
\newcommand{\airi}{\text{\textdaggerdbl}}
\newcommand{\sber}{\text{\textdollar}}
\author{Oleg Serikov$^{\airi\mipt\hse}$, \quad Vitaly Protasov$^{\airi}$, \quad Ekaterina Voloshina$^{\sber}$, \quad Viktoria Knyazkova$^{\hse}$, \\ \textbf{Tatiana Shavrina}$^{\airi \sber}$\medskip\\
         $^{\airi}$ Artificial Intelligence Research Institute, \quad $^{\sber}$ SberDevices,\\
         $^{\hse}$ HSE University, \quad $^{\mipt}$ DeepPavlov lab, MIPT}

\begin{document}
\maketitle
\begin{abstract}

Linguistic analysis of language models is one of the ways to explain and describe their reasoning, weaknesses, and limitations. 
In the probing part of the model interpretability research, studies concern individual languages as well as individual linguistic structures. 
The question arises: are the detected regularities linguistically coherent, or on the contrary, do they dissonate at the typological scale? Moreover, the majority of studies address the inherent set of languages and linguistic structures, leaving the actual typological diversity knowledge out of scope.
In this paper, we present and apply the GUI-assisted framework allowing us to easily probe a massive number of languages for all the morphosyntactic features present in the Universal Dependencies data. We show that reflecting the anglo-centric trend in NLP over the past years, most of the regularities revealed in the mBERT model are typical for the western-European languages. 
Our framework can be integrated with the existing probing toolboxes, model cards, and leaderboards, allowing practitioners to use and share their standard probing methods to interpret multilingual models.
Thus we propose a toolkit to systematize the multilingual flaws in multilingual models, providing a reproducible experimental setup for 104 languages and 80 morphosyntactic features. \href{https://github.com/AIRI-Institute/Probing_framework}{GitHub}




\end{abstract}

\section{Introduction}
Probing methods shed light on the black box of the neural models in unearthing the linguistic features encoded in them. Probing sets a standard setup with various internal representations from the model and uses an auxiliary classifier to predict linguistic information captured in the representation. 

As probing research results have come up with contradictory results on different languages and language models, there appears to be a methodological need for a meta-study of the accumulated knowledge and a need to standardize the experimental setup. At the same time, the fixation of the setup and hyperparameters should allow the reproduction of a wide range of experiments, such as multilingual probing, like X-Probe \cite{Ravishankar2019ProbingMS} and Linspector \cite{Sahin2020LINSPECTORMP}, layer-wise probing \cite{fayyaz-etal-2021-models}, chronological probing \cite{voloshinaetal}.

Often, data for probing experiments is based on already known competition data, benchmarks, and gold standards. To obtain consistent results, such data must be high-quality, manually validated, and carefully include multiple languages.
For this reason, in this work, we use the Universal Dependencies data \cite{de-marneffe-etal-2021-universal} as a source of multilingual data with a validated and standardized complete morphological and syntactic annotation, which will allow us to accumulate the assimilation of specific linguistic phenomena in many languages at once.
Probing these languages on the respective annotated linguistic categories would reveal how models seize the typological proximity of languages.

Therefore, the general probing methodology should include (according to \citet{Conneau2018SentEvalAE}) 1) a fixed set of evaluations based on what appears to be community consensus; 2) a fixed evaluation pipeline with standard
hyperparameters; 3) a straightforward Python interface. 

This paper aims to extrapolate the multilingual linguistic diversity on the proven and tested SentEval-like methodology.

We state our contribution as follows:
\begin{itemize}
    \vspace{-0.2cm}\item We develop a framework for exhaustive multilingual probing of the language models, with a complete enumeration of all grammatical characteristics and all languages available in Universal Dependencies while maintaining the standard SentEval format.
    \vspace{-0.2cm}\item We provide a setup for better and explanatory aggregation and exploration of the massive probing results with thousands of experiments for each model.
    \vspace{-0.2cm}\item We illustrate the possibilities of the framework on the example of the mBERT model, demonstrating new insights and reassuring the results of previous studies on narrower data.
\end{itemize}


Performing probing studies on such a large scale addresses the vision outlined in \citet{nichols-2007} and contribute to a new dimension to linguistic typology research, as the revealed structures are encapsulated in tools and data inseparably tied to nowadays linguistic nature. 
Our framework provides users from different fields, including linguists, with a new point of view on the typological proximity of languages and categories.

\section{Related Work}


Different attempts were made to interpret behavior and hidden learned representation of language models. For example, \citet{Hoover2020exBERTAV} investigated the attention-heads of the BERT model on word tokens connectivity level. \citet{Wallace2019AllenNLPIA} presented an interpretation framework where they improved a visual component of the model prediction process on several NLP tasks for the end-user.

Flourishing after the ACL debates on semantic parsing\footnote{\url{https://aclanthology.org/volumes/W14-24/}}, the probing methodology has developed its own model interpretation tools. 
Thus, \textbf{SentEval framework} \cite{Conneau2018SentEvalAE} includes various types of linguistically-motivated tasks: surface tasks probe for sentence length (SentLen) and for the presence of words in the sentence (WC); syntactic tasks test for sensitivity to word order (BShift), the depth of the syntactic tree (TreeDepth) and the sequence of top-level constituents in the syntax tree (TopConst); semantic tasks check for the tense (Tense), the subject (resp. direct object) number in the main
clause (SubjNum, resp. ObjNum), the sensitivity to random replacement of a noun/verb (SOMO) and the random swapping of coordinated clausal conjuncts (CoordInv).

\textbf{Linspector}  \cite{https://doi.org/10.48550/arxiv.1903.09442} includes 15 probing tasks for 24 languages by taking morphosyntactic language properties into account, including case, verb mood, and tense, syntactic correctness, and the semantic impossibility of an example. While lacking the simplicity of the SentEval approach, the framework provides both a linguistically-grounded and multilingual setup. We are significantly expanding both the list of languages and properties being examined.

\textbf{Probe-X} \cite{ravishankar-etal-2019-probing} has expanded SentEval setup with 5 additional languages, while \textbf{NeuroX framework} \cite{neurox-aaai19:demo} also introduced novelty, but proposed to enrich the methodology to allow for cross-model analysis of the results, supporting neuron-level inspection.

\subsection{Probing Critique}
We would state a few problems why some of the probing practices are methodologically problematic. 

First, the probing interpretation result can differ from paper to paper, creating various conclusions from different authors. While \citet{jawahar-etal-2019-bert} achieves from 69.5-96.2\% accuracy on the SentLen SentEval probing task (BERT model), they state that this info is somehow represented at the bottom layers.
The work \cite{ravishankar-etal-2019-probing} achieves 38-51\% accuracy on SentLen (RNN encoder) and states that "recurrent encoders show solid performance on certain tasks, such as sentence length."
This drastic difference in result interpretation (``somehow'' vs. ``extremely strong'') leads to misrepresenting the factual results. Conflicting evidence within the field of BERTology can be found in \citet{rogers-etal-2020-primer}, see Sec 3.1 and 4.3.

Secondly, the results on similar tasks can be obtained with unstable success if the hyperparameters are not fixed or exhaustively described: for example, study \cite{jawahar-etal-2019-bert} finds that "BERT's intermediate layers encode a rich hierarchy of linguistic information, with surface features at the bottom, syntactic features in the middle and semantic features at the top," while the work by \citet{tikhonova_mikhailov_pisarevskaya_malykh_shavrina_2022} on mBERT shows, that the model does not learn the linguistic information.
More meta-research is needed to explore the contradictory results obtained by the community.

\subsection{Task Representation}
In the survey of post-hoc language model interpretation \cite{Madsen2021PosthocIF}, the linguistic information-based tasks fall into the groups of the highest abstraction and the top-informativeness of properties used. 
This group of projects includes tasks based on the various theoretical language levels: from part-of-speech tagging to discourse. 
\paragraph{Languages}
While the most tasks are English-based, there appear the non-English monolingual frameworks: French-based probing \cite{merlo-2019-probing}, Russian-based SentEval \cite{mikhailov-etal-2021-rusenteval}, Chinese word masking probing \cite{Cui_2021}.
The multilingual benchmarks have paved the way for multilingual probing studies by collecting the necessary data.

\paragraph{Linguistic features}
Most language-based tasks tend to be based on morphology or syntax, deriving from SentEval methodology. Thus, higher-level tasks can concentrate both on monolingual discourse evaluation \cite{Koto2021DiscoursePO} (mostly English-based by now), as well as the multilingual discursive probing based on the conversion of the existing multilingual benchmarks \cite{kurfali-ostling-2021-probing} (XNLI, XQUAD). 

\section{Framework Design}

This section describes the probing framework and the experimental setup part.

The main goal is to probe how well a model assimilates language constructions during training. For the framework, we want to form an end-to-end solution that can be applied to different models, work on diverse data, and simplify the process of getting insights from the results.

Based on that, the challenges we have are the following: 
\begin{enumerate}
    \vspace{-0.3cm}\item The data we use in the training and evaluation parts must be in the standard format no matter what language we deal with.
    \vspace{-0.3cm}\item The probing process should be universal for different models. Based on it, we also need to collect detailed results for further analysis.
    \vspace{-0.3cm}\item Since we aim to work with diverse data, we should contain instruments to simplify the process of getting insights from the results. If we do not handle this problem, we can have bunches of results that would be difficult to interpret and provide findings for.
\end{enumerate}

Thus, we can represent our framework as a tool with different instruments. The first one is aimed at pre-processing data for probing, which is commonly a classification task. The second one is a probing engine supporting popular probing techniques such as diagnostic classification. And the last one is a visualization instrument which should ease the process of interpreting the findings.

\subsection{SentEval Format Converter}
We found the SentEval format to be generally good and universal in the data composition for classification tasks. Since we have such a vast resource as Universal Dependencies for different languages, we can transform the data into the SentEval format and compose different classification tasks based on the language categories we can get.

\begin{figure*}[!h]
    \centering
    \includegraphics[width=\textwidth]{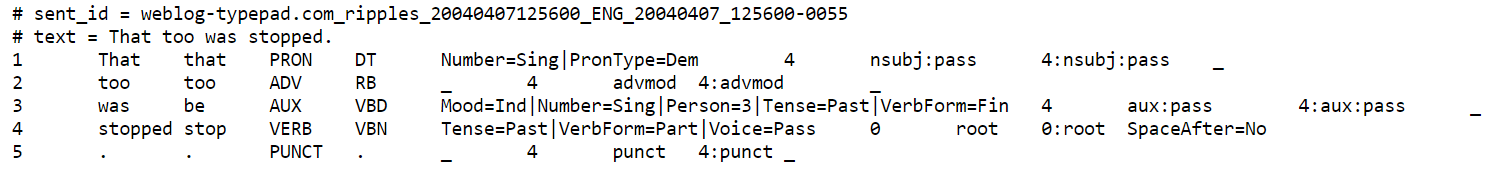}
    \caption{The example of UD annotation}
    \label{fig:ud_example}
\end{figure*}

UD annotation consists of several parts: lemmas, parts of speech, morphological features, and universal dependencies relations. The converter to SentEval format is focused on morphological features. As Table \ref{fig:ud_example} illustrates, morphological categories are written in the sixth column with their category values separated by the equals sign, for example, in \textit{Number=Sing}, \textit{Number} is a category and \textit{Sing} is a category value.
It took us 8 hours to process by the SentEval converter on 96 CPUs for absolutely all archives.



\begin{table*}[h]

\begin{tabular}{p{1.3cm}l }
\textbf{Format} & \textbf{Data entry}
\\
    \textit{Conll-U} & 
\begin{lstlisting}[basicstyle=\small, mathescape=true]
# sent_id = weblog-typepad.com_ripples_20040407125600_ENG_20040407_125
# text = That too was stopped.
1. $\textbf{That}$  that PRON DT Number=Sing|PronType=Dem 4 nsubj:pass 4:nsubj:pass _
2. $\textbf{too}$  too ADV RB _ 4 advmod 4:advmod _
3. $\textbf{was}$  be AUX VBD Mood=Ind|Number=Sing|Person=3|Tense=Past|VerbForm=Fin 4 
aux:pass 4:aux:pass  _
4. $\textbf{stopped}$  stop VERB VBN Tense=Past|VerbForm=Part|Voice=Pass 0 root 0:root 
SpaceAfter=No
5. $\textbf{.}$  .  PUNCT  .  _  4  punct  4:punct  _
\end{lstlisting}
\medskip
\\
\textit{SentEval} & 
\begin{lstlisting}[basicstyle=\small]
tr Past That too was stopped .
\end{lstlisting}

\end{tabular}
\caption{Example of CONLL-U format and its conversion to SentEval: Tense classification, train set.}
\label{fig:ud_annottion}


\end{table*}

For each morphological category found in a given file, the converter generates a new file in SentEval format according to the following steps:

\begin{algorithm}[]
 \KwData{CONLLU files or a directory to such files for one language}
 \KwResult{a file in SentEval format}
 read files\;
 find all morphological categories\;
 \ForEach{categories}{
  \ForEach{sentences}{
  \uIf{category is in sentence}{
    get a category value}}
  stratified split on three samples\;
  write to a file
 }
 \caption{The conversion process}
\end{algorithm}

If split UD data into train, validation, and test sets, we do not change this split. In other cases, we split data into three sets, so the distribution of category values in the original text will be kept in each set.


If a sentence contains several words with the same morphological categories, the closest to the sentence node word is taken, preventing the one sentence from being repeated several times. Table \ref{fig:ud_annottion} depicts the example of \textit{Tense} category, the value of word \textit{stopped} will be taken, as it is the root of the sentence.

\subsection{Multilingual Data} 
\label{sec:data}
We take 289 repositories, including the data of 172 languages available at the GitHub of Universal Dependencies , updated in May 2022.\footnote{\url{https://github.com/UniversalDependencies}}

While parsing files, we face several problems inherited from UD. 71 of the repositories do not contain any CONLLU files. Three Japanese repositories and Korean and Frisian Dutch repositories contain different annotations from standard UD annotations. The data from 16 repositories (Akkadian, Cantonese, Chinese (2), German, Japanese, Hindi, Irish, Kangri,  Maltese, Neapolitan, South Levantine Arabic, Swedish Sign language, Swiss German, Old Turkish, Tagalog) do not contain morphological annotation. 
Also, some repositories include correctly annotated data but are not suitable for classification problems because all the examples contain only one value of all the categories, for example, only examples with class \textit{Plural} are left for the category Number (Cantonese, Chukchi, Frisian Dutch, Hindi English, Japanese, Kangri, Khunsari, Makurap, Maltese, Nayini, Neapolitan, Old Turkish, Soi, South Levantine Arabic, Swedish Sign Language, Swiss German, Telugu, Vietnamese).

After filtering, we have data from 104 languages from 194 repositories (see Appendix \ref{sec:languages}). From the typological point of view, these languages belong to 20 language families, and the Basque language is an isolate. Although almost half of the languages are from the Indo-European family, the data include several under-studied language families. Many of the languages in our data are endangered or even extinct.
The UD data is distributed based on Creative Commons and GNU-based licenses, varying from language to language\footnote{\url{https://lindat.mff.cuni.cz/repository/xmlui/page/licence-UD-2.1}}.
Extracting the tasks for every grammatical category results in 1927 probing datasets.
\subsection{Probing Engine}

\subsubsection{Encoders}
In the experiments, we consider the layers of encoder-based models and their ability to acquire language data and perform well on probing tasks.
Using the output of the model's layers, we can get contextualized token embeddings for elements of the input text. For that reason, we can consider several options for embedding aggregation: \textbf{CLS} where the text is presented as the embedding from "[CLS] "token, \textbf{SUM} and \textbf{AVG} where the sentence vector is a sum or average of embeddings of all text tokens.

\subsubsection{Classifiers and metrics}
After the embeddings are obtained, we train a simple classification model based on the encoder layers' representation and task data labels. We consider linear (Logistic Regression) and non-linear (MLP) classifiers.
As the metrics for performance evaluation, we use \textit{accuracy} score and weighted $F_{1}$ score in case of unbalanced classes.

\subsection{Aggregation}
The engine is meant to produce probes of a particular category in a particular language. We provide additional grouping and visualization tools to allow for meaningful interpretation of such large result sets.
They are meant to highlight similar experiments and visualize them on the world map.
 
The default configuration follows the classical probing experiments and uses layers' numbers as X axes. Yet the more novel setup can be chosen, e.g. treating the $<\text{language}, \text{category}>$ features pairs as X-axis instead.

The defined atomic experimental axis allows to characterize larger groups of experiments via their pooled value (such as mean-pooled by categories value in Figure \ref{fig:vitalya_layers}), or even cluster them (e.g., using pairwise experiments similarity as in Figure \ref{fig:gui}). 


\subsubsection{Similarity Metrics}
We support two metrics of scoring the experiments' pair-wise similarity. Both of them are calculated for the experiment results curves. \footnote{By probing curve we refer to the typical probing chart. Layers, or other probed parts of a model, and the respective results are visualized as a curve on a linear chart.}
\textit{Frechet distance} (\frechet) provides a natural way to compare curves taking into account both the similarity of curves' shapes and their absolute positioning on the chart.
Unlike that, for \textit{Pearson correlation} (\pearson) absolute positioning is irrelevant.

\begin{figure}
    \centering
    \includegraphics[width=\linewidth]{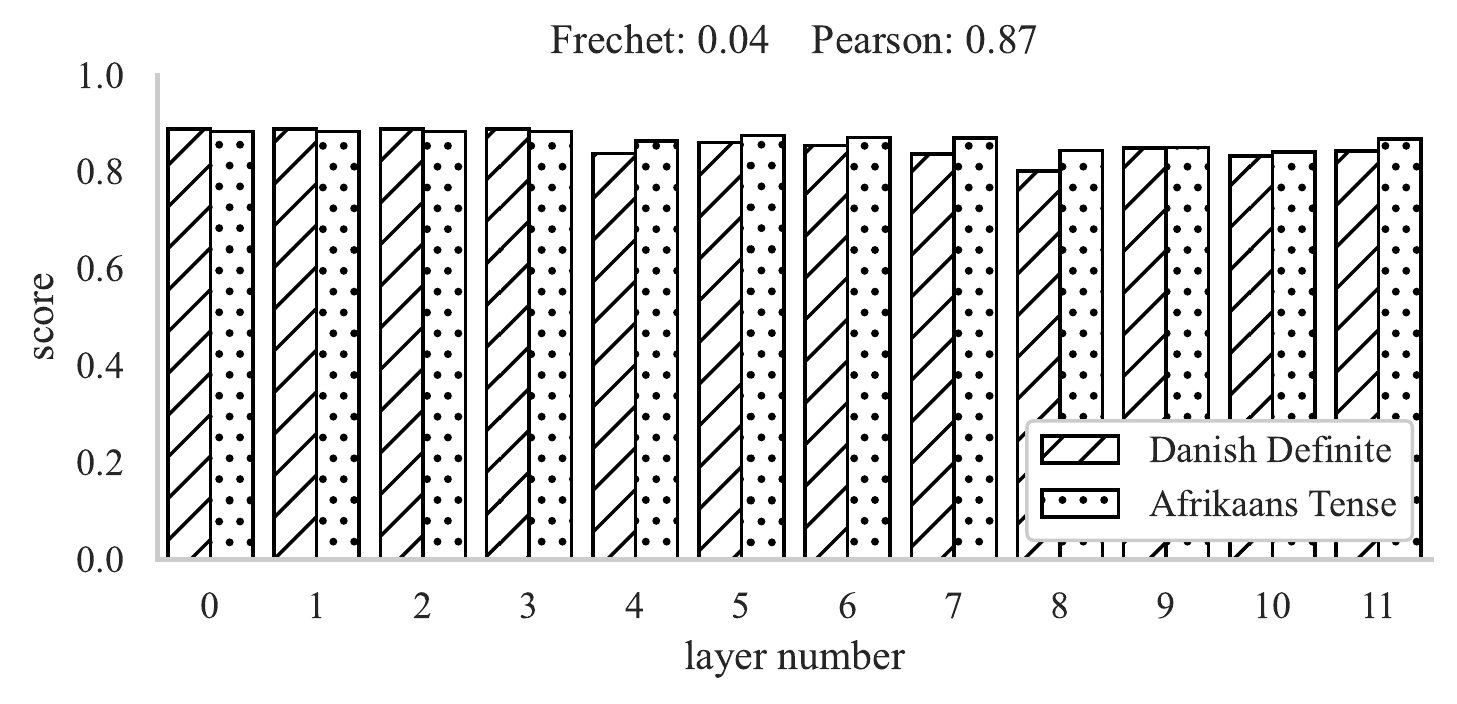}
    \caption{An example of \frechet and \pearson scores calculation between the two probing experiments}
    \label{fig:sim_example}
\end{figure}

While \pearson formalizes the notion of ``coherent'' or ``similar'' behavior of models' layers, \frechet complements it with exact values similarity constraint (see Figure \ref{fig:sim_example}).

\paragraph{Frechet distance}

Given the simultaneous iterative step-by-step walkthrough from the start to the end points of both curves, one could freely vary the step size for every curve at every iteration. 
By the proper choice of step sizes during the walkthrough, one could guarantee that the optimal distance between curves' respective points will never be exceeded during the iteration process.
That optimal distance is called Frechet distance and is formally calculated as follows: 
$\delta_F = \text{inf}_{a,b} \left\{   \text{max}_t \left\{ d\left( A_a(t), B_b(t) \right) \right\} \right\}$, where $t$ denotes iteration steps, $a,b$ combinations correspond to various step size strategies, and $A, B$ are the functions respective to the curves.  

\paragraph{Pearson correlation coefficient}
Pearson correlation measures the strength of linear dependence of two samples: 
$r_{xy} =\frac{\sum x_iy_i-n \bar{x} \bar{y}}{(n-1) s_x s_y}$, where $s_\alpha$ is the standard deviation of sample $\alpha$ and $\overline{\alpha}$ is this sample mean.

\subsubsection{Visualization}
\begin{figure*}[htb]
    \centering
    \includegraphics[height=7cm, ]{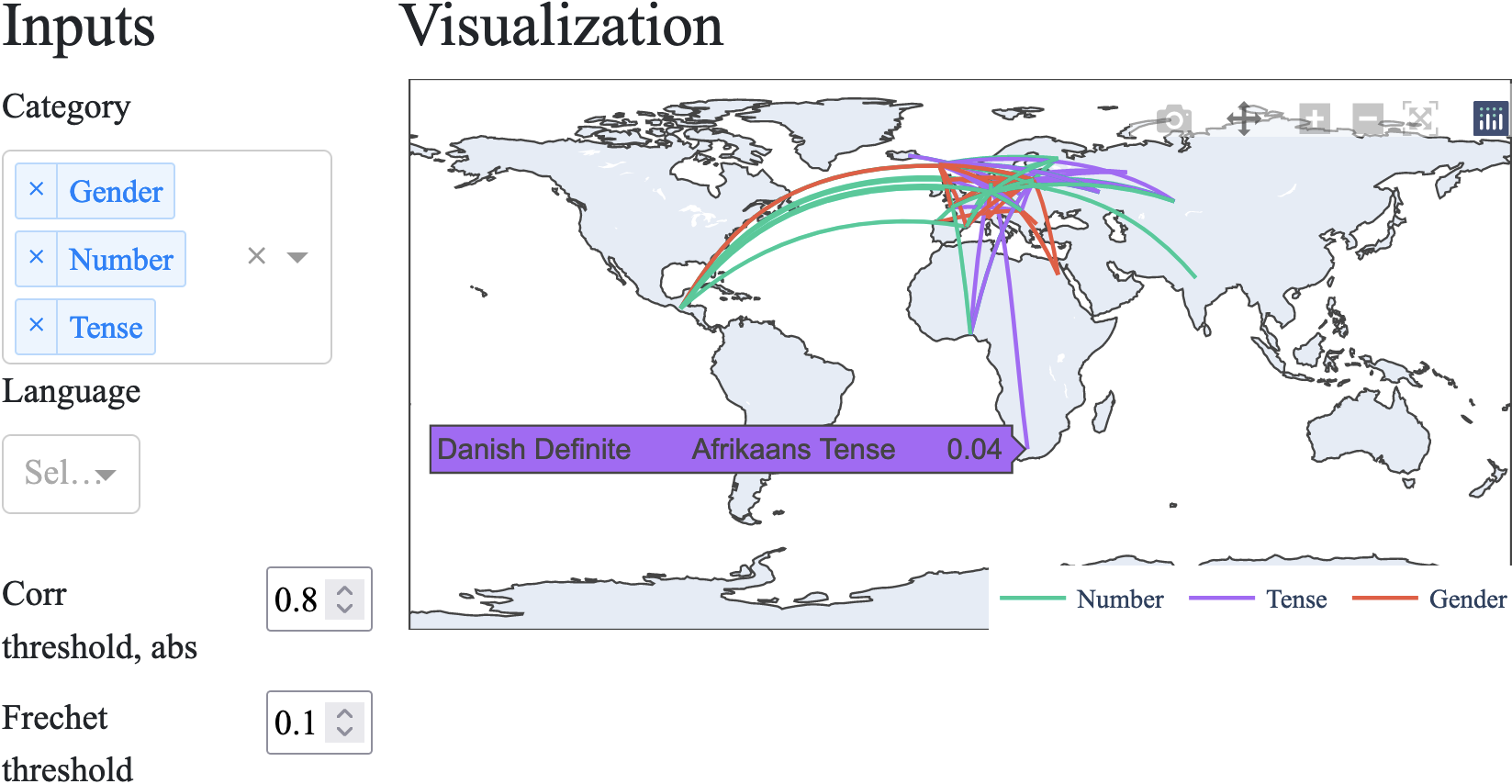}
    \caption{GUI screenshot: Similarity between languages learned by mBERT based on different probing tasks.}
    \label{fig:gui}
\end{figure*}

We provide the GUI (see Figure \ref{fig:gui}) to allow us to configure the similarity thresholds and explore the particular categories' results on a geospatial chart. 

GUI allows setting off the \frechet and \pearson absolute values thresholds and specifying particular languages and categories to be shown.


\section{Evaluation Setup}
To present the whole procedure of our probing framework working process, we decided to run the experiments only for two multilingual transformer encoder-based models: the 12-layer mBERT model \cite{https://doi.org/10.48550/arxiv.1810.04805}\footnote{\url{https://huggingface.co/bert-base-multilingual-cased}} and the 24-layer XLM-RoBERTa model \cite{DBLP:journals/corr/abs-1911-02116}\footnote{\url{https://huggingface.co/xlm-roberta-large}}. We used embeddings from “[CLS]“ token for each text sample as it is widely accepted.
As the classifier, while supporting LogReg and MLP, we choose Logistic Regression due to its higher \textit{Selectivity} \cite{Hewitt2019DesigningAI}. 
The classifier was trained on 10 epochs using cross-entropy loss and \textit{AdamW} \cite{DBLP:journals/corr/abs-1711-05101} optimizer. A separate classifier was trained for each feature of all languages and each layer.

To eliminate the problem of different sizes of the datasets, we run the classifier five times and then take an average result to avoid the classifier bias. The results were evaluated by $F_1$ weighted score because of the unbalanced data for most probing tasks. From Universal Dependencies, using our SentEval converter, we obtained 1927 probing tasks for 104 languages.
During the training, we noticed that some samples contain long sentences with token numbers of more than 512. We propose two options for handling it correctly: truncate sentences to 512 tokens or dispose of all of these sentences.



\section{Results and Insights}
\subsection{General Results}
We received a massive multilingual probing task bundle of 1927 tasks using all the converted data for 104 languages. It took us 10 hours to probe through all the files on one NVidia Tesla GPU V100.

\begin{figure*}[!htb]
    \centering
    \includegraphics[height=14cm,]{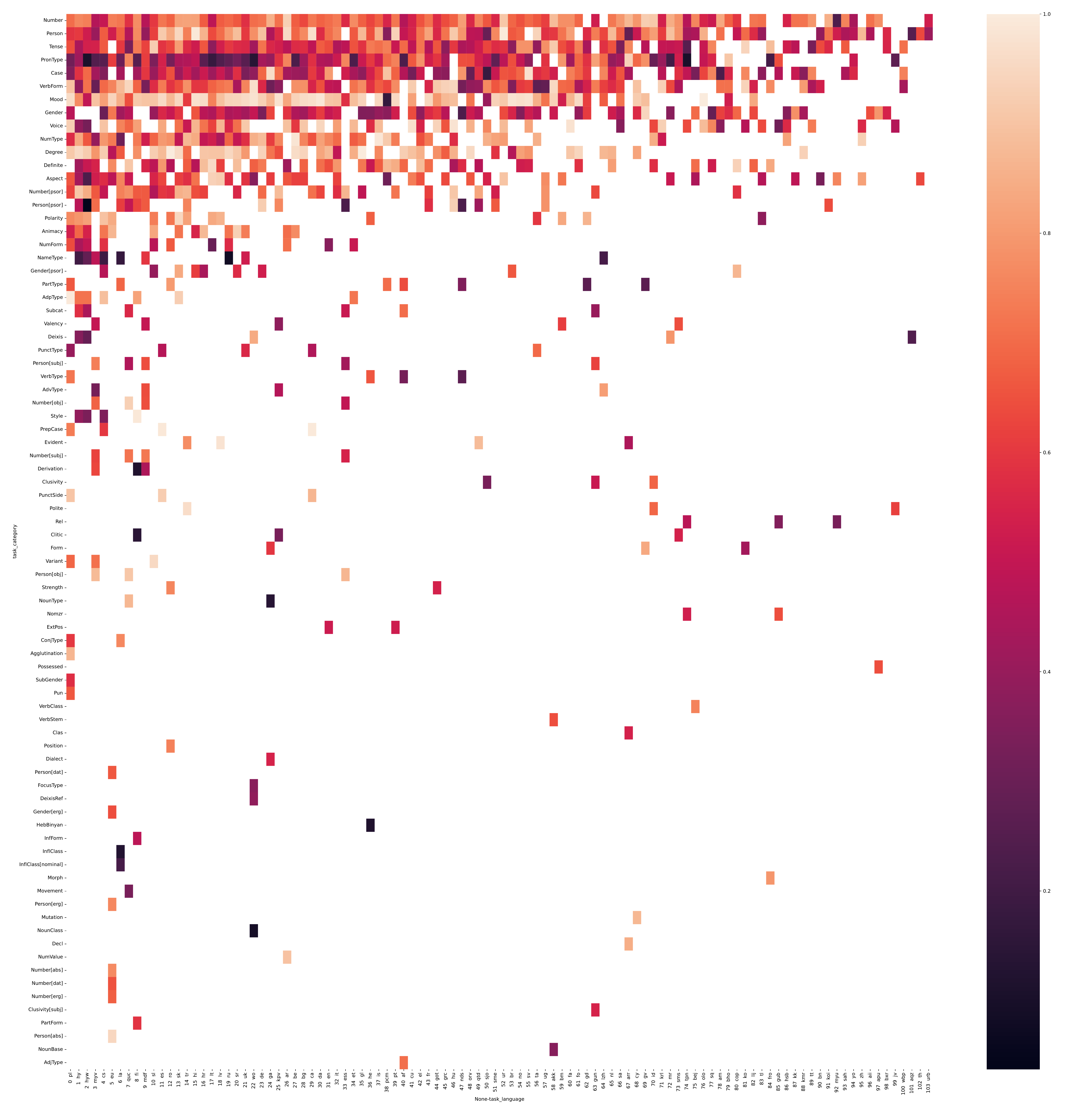}
    \caption{mBERT results grouped by languages and average feature probing score on all layers}
    \label{fig:oleg_hmap}
\end{figure*}

\begin{figure}[!htbp]
    \centering
    \includegraphics[width=\linewidth]{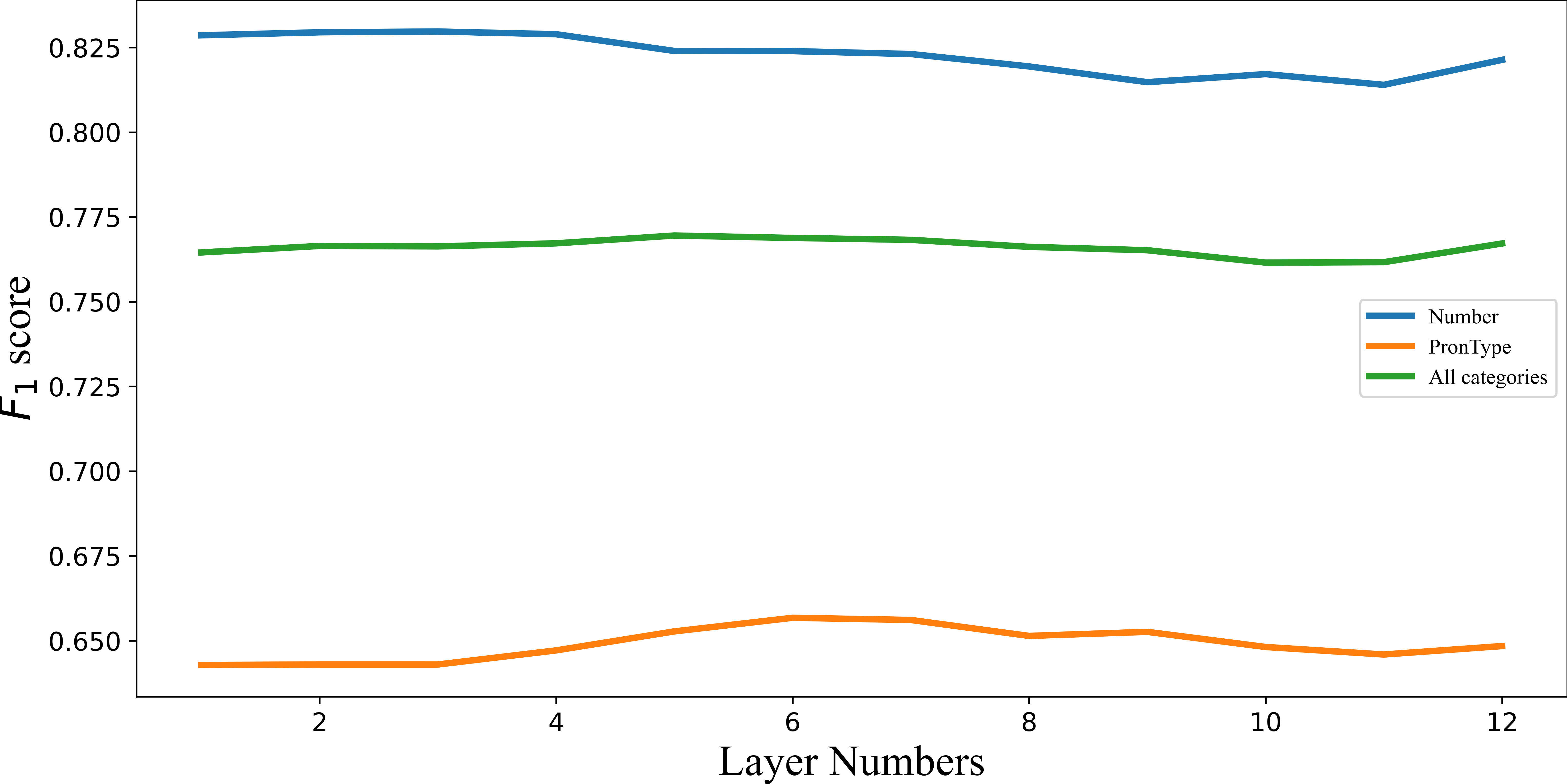}
    \caption{Distribution of scores by model's layers depending on the categories which are used across languages.}
    \label{fig:layers_res}
\end{figure}

We thus conducted the probing of the mBERT and XLM-R models to figure out the capabilities of the models, as follows:
\begin{enumerate}[topsep=0pt,itemsep=0pt,partopsep=0pt, parsep=0pt]
    \item to generalize linguistic information language-wise:  grouping the average results a) by layers (Figure~\ref{fig:layers_res}), b) by each feature in each language (Figure~\ref{fig:oleg_hmap}).
    \item to generalize linguistic information feature-wise, grouping the average results by each layer and by each language (Appendix~\ref{sec:heatmap_mbert}).
    \item to explore the results feature-wise: a) by searching for similarities in layer-wise feature representations) by exploring individual feature results grouped by language and layer (Appendices \ref{sec:heatmap_number}, \ref{sec:heatmap_prontype}), c) by creating the geospatial visualizations of the similar features. 
\end{enumerate}
The model evaluation results are presented in Figure~\ref{fig:oleg_hmap}: the figure clearly shows the sparseness with which all features are presented in each language. Basic features such as Number, PronType, and Tense are among the most frequent ones. 
The example of the geospatial visualizations of the similarly learned features is presented in Figure~\ref{fig:gui}



\subsection{mBERT and XLM-R Multilingual Abilities and Insights}
Given the mBERT model as an example, as for the categories \textit{Number} and \textit{PronType}, which are the most common across languages, the best scores were achieved at the 3rd and 6th layers, respectively. In the case of all categories, mBERT showed the best result at the 5th layer.
In Appendix \ref{sec:heatmap_number} and \ref{sec:heatmap_prontype} the heatmaps for all languages with these categories can  be found. As for the average results by all categories, see Appendices \ref{sec:heatmap_mbert} and \ref{sec:heatmap_xlm_r}. Figure \ref{fig:layers_res} depicts average language scores for Number and PronType and among all categories. 

As mentioned above, we evaluated two models, mBERT and XLM-R, on our data. On average, their performance is similar. However, the scores of XLM-R are slightly worse than the ones of mBERT.
By the categories, mBERT shows the best performance in Hungarian, Chinese, Urdu, Welsh, and Slovak. The worst results were shown in Tupinambá and Akuntsu. XLM-R's top languages include the same language in a different order (Chinese has the best quality).

On Javanese and Akuntsu, XLM-R shows the worst quality. The models show the best quality on high-resource languages and have worse representations of under-resourced non-Indo-European languages. 

Among the factors that can impact the models' performance on different languages, the following might be the most essential: script, language genealogy, and typological features of languages. 

To research the effect of script and language genealogy on the results, we run an ANOVA test since we have more than two families or scripts. The test reveals a strong correlation between a language family and the models' performance ($p = .0005$ for both XLM-R and mBERT). 

We also run an ANOVA test to see if another significant difference in performance on languages with different scripts exists. The test shows that script did not impact the final performance ($p = .52$ for mBERT and $p = .39$ for XLM-R). The reported difference in the performance may be caused by the set of categories or the dataset size. Independently, other works \citep{pires2019multilingual, wu2019beto} claim that mBERT shows language neutrality regarding both a language and its script. Our results support that level of performance does not depend on a script, as models show high results on languages with Arabic-based (Persian, Urdu) or Ge'ez scripts (Amharic).

Yet, the models might be biased towards Standard Average European languages (SAE) \citep{haspelmath2001sae}, as it solves tasks on the categories found in SAE languages better than on the other language-specific categories. For example, the top-10 of the best recognised categories include \textit{Person[abs], PunctSide, Person[obj], Agglutination \footnote{Only used for Polish past participles}, Mutation, Degree, Decl, Mood \footnote{Mood express modality, such as indicative, imperative, conditional}, Evident\footnote{Evidence is the morphological marking of a speaker’s source of information \citep{aikhenvald2006evidentiality}}}, and \textit{Polarity\footnote{Polarity shows if words can be used only in negative or positive contexts}}. Agglutination and Mutation are highly imbalanced towards one class, and other categories, except for Evident, are widespread in European languages. 

On the other hand, top-10 worst categories are following: \textit{NumValue, InflClass, NounClass, HebBinyan, Clitic, ExtPos, Derivation, NameType, InflClass[nominal], FocusType}. These categories are language-specific and are not found in SAE languages. Apart from that, if a category typical for SAE gets a different set of values, the model performs much worse. The model generally shows good results on \textit{Case} and significantly worse results for Hungarian and Amharic with a different set of values for \textit{Case}.

\citet{chi2020finding} prove that mBERT has a joint subspace of universal syntactic relations. Since we cannot fully prove if mBERT has a joint subspace of morphological features because, as mentioned before, some morphological features are not universal. However, we can see if there is any correlation between categories across all languages based on how different layers learn these features, according to Frechet distance. Most categories do not have a correlating category. However, there are several compelling cases to mention. mBERT generally learns Evident and Mood similarly, and in some languages, such as Bulgarian, Evident is regarded as a value of \textit{Mood}. Other than that, Definite and Number have a little distance, which might be expressed with one morpheme, as in Scandinavian languages. The same is valid for Number and Person categories that are learned similarly by mBERT.

There is not enough evidence to claim that mBERT has a joint subspace for morphological features because categories have different sets of values, and mBERT performs better on SAE categories. Yet, it shows some generalization abilities on the similarity of morphology across languages.




\section{Future Work}
As part of future work, we plan to include syntactic markup in research and an interface with a CQL-like\footnote{\url{https://www.sketchengine.eu/documentation/corpus-querying/}} query language for authors. The ability to set conditions on a subcorpus of examples will give the researchers the freedom to create custom and linguistically motivated probing tasks while the rest of the experiment parameters will be fixed. One can imagine, e.g., the probing of the model on the \verb|[tag="NP"]| query, exploring the results specifically on the noun phrases. We believe that, in this respect, the tasks of probing and interpreting the results of the model become close to the tasks of corpus linguistics and searching through corpora for statistical testing of hypotheses.



\section{Conclusion}
The typological variety of linguistic features composes the general nature of language. To address the lower-abstract parts of this nature, we introduced the Universal Probing framework, which allows the researchers to run and aggregate massive amounts of probing experiments in a fixed and reproducible setup.

The current framework version includes an experimental setup from 104 languages and 80 grammatical features.
The framework can be used for language model interpretation with various architectures, and the results can also be easily incorporated into the model cards checklist. It can be used in more language-wise transfer learning and typological studies with multilingual models.

We hope that the community will use our work in order to interpret, evaluate and compare the language models, leading to better and more explainable NLP.
The framework and all the data are open-source under Apache 2.0 license \url{https://github.com/AIRI-Institute/Probing_framework}. \newline \newline

\section{Limitations}
\label{sec:limit}
By now, it is also worth mentioning the UD data dependencies of the framework. The problems in the UD data, such as annotation errors, formatting errors, and version instability, could potentially affect the resulting probing framework. As described in Section~\ref{sec:data}, we have eliminated obviously problematic fragments; however, more deeply incorporated inaccuracies may drag on, surviving conversion to the SentEval format. Some inconsistencies in the accepted annotation format affected the quality of model embeddings: such categories as PunctSide (Catalan, Finnish, Icelandic, Polish, Spanish), NameType (Armenian, Classical Chinese, Czech, Erzya, etc.) are rare and have a very different distribution from language to language and are expected to be at the bottom of the list. The categories accepted in the UD for one specific language are also poorly solved: Agglutination (Polish), Mutation (Welsh), HebBinyan (Hebrew), NounClass (Wolof), NumValue (Arabic, Czech).

We use the latest available UD release (version 2.10)\footnote{Version 2.10 treebanks are available at \url{http://hdl.handle.net/11234/1-4758}. 228 treebanks, 130 languages, released May 15, 2022.}. As stated on the project's website, the next release (v2.11) is scheduled for November 15, 2022, so data curation and updates will be necessary to incorporate the newer and better UD annotation into the framework.

The proposed framework allows for different probing methods to be used similarly, including the widely criticized ones \cite{belinkov2022probing}. Researchers relying on the presented framework should carefully pick the proper methods in their probing studies. For example, we've introduced the control task \cite{Hewitt2019DesigningAI} consisting of averaging the probing performance across several probing experiments. This reduced the possibility of a probing task erroneously receiving a high score due to the small size of the testing data.

\section{Ethical Considerations}
\subsection{Possible Misuse}
The framework's usage implies working concerning standard practices during model pre-training, such as controlling that the test data (e.g., UD corpora) are excluded from the training corpus. Using UD data during pre-training or fine-tuning the model can lead to indicative and biased results of model interpretation.

\subsection{Data-specific Problems}
\subsubsection{Dataset Characteristics}
The dataset covers the languages described in Section~\ref{sec:languages}. The probing dataset statistics are also presented in Section~\ref{sec:languages}.

\subsubsection{Generalization}
The UD data can be considered mostly validated, as it involves multiple institutions to develop and test the annotation standards, as well as the corpus data itself. However, besides data quality, usage of the data should address such characteristics as quantity: that is why we have automatically excluded the UD categories having only one value within a category in all available languages. For all other categories, the data for the classification task were not limited in any way; the train/val/test data division was preserved. 

Potentially, other data-dependent problems (see also the resulting data dependencies in Section~\ref{sec:limit}) could be:
\vspace{-3mm}
\begin{itemize}[itemsep=0pt]
    \item genre bias in specific languages;
    \vspace{-2mm}\item personal style/resource bias in specific languages;
    \vspace{-2mm}\item collocation of the specific features: some features can possibly occur within the same contexts (sentences), which makes the solution of the classification problem within the probing setup noisy for the tested language model.
\end{itemize}

\vspace{-3mm}

Nevertheless, we consider the UD corpora to be sufficiently reliable and the most complete of the available data for a detailed low-level multilingual probing study of the models.

\subsubsection{Data Quality}
In addition to the above, we draw attention to the question of the language representation problems in the UD. According to the Ethnologue database \footnote{\url{https://www.ethnologue.com/enterprise-faq/how-many-languages-world-are-unwritten-0}}, there are more than 4000 languages with developed writing systems, while only 172 of them are presented in the UD in general, and even less (104) were qualified to be included in the framework format.

As we understand that the presented language set is not typologically sampled, we proceeded from the criterion of completeness, not balance. If necessary, we encourage willing researchers to sample their subsamples from our data to follow typological sampling.

\bibliography{anthology,custom}

\begin{thebibliography}{31}
\expandafter\ifx\csname natexlab\endcsname\relax\def\natexlab#1{#1}\fi

\bibitem[{Aikhenvald(2006)}]{aikhenvald2006evidentiality}
Alexandra~Y Aikhenvald. 2006.
\newblock Evidentiality. oxford: Oxford university press, 2004.

\bibitem[{Belinkov(2022)}]{belinkov2022probing}
Yonatan Belinkov. 2022.
\newblock Probing classifiers: Promises, shortcomings, and advances.
\newblock \emph{Computational Linguistics}, 48(1):207--219.

\bibitem[{Chi et~al.(2020)Chi, Hewitt, and Manning}]{chi2020finding}
Ethan~A Chi, John Hewitt, and Christopher~D Manning. 2020.
\newblock Finding universal grammatical relations in multilingual bert.
\newblock \emph{arXiv preprint arXiv:2005.04511}.

\bibitem[{Conneau et~al.(2019)Conneau, Khandelwal, Goyal, Chaudhary, Wenzek,
  Guzm{\'{a}}n, Grave, Ott, Zettlemoyer, and
  Stoyanov}]{DBLP:journals/corr/abs-1911-02116}
Alexis Conneau, Kartikay Khandelwal, Naman Goyal, Vishrav Chaudhary, Guillaume
  Wenzek, Francisco Guzm{\'{a}}n, Edouard Grave, Myle Ott, Luke Zettlemoyer,
  and Veselin Stoyanov. 2019.
\newblock \href {http://arxiv.org/abs/1911.02116} {Unsupervised cross-lingual
  representation learning at scale}.
\newblock \emph{CoRR}, abs/1911.02116.

\bibitem[{Conneau and Kiela(2018)}]{Conneau2018SentEvalAE}
Alexis Conneau and Douwe Kiela. 2018.
\newblock Senteval: An evaluation toolkit for universal sentence
  representations.
\newblock \emph{ArXiv}, abs/1803.05449.

\bibitem[{Cui et~al.(2021)Cui, Che, Liu, Qin, and Yang}]{Cui_2021}
Yiming Cui, Wanxiang Che, Ting Liu, Bing Qin, and Ziqing Yang. 2021.
\newblock \href {https://doi.org/10.1109/taslp.2021.3124365} {Pre-training with
  whole word masking for chinese {BERT}}.
\newblock \emph{{IEEE}/{ACM} Transactions on Audio, Speech, and Language
  Processing}, 29:3504--3514.

\bibitem[{Dalvi et~al.(2019)Dalvi, Nortonsmith, Bau, Belinkov, Sajjad, Durrani,
  and Glass}]{neurox-aaai19:demo}
Fahim Dalvi, Avery Nortonsmith, D.~Anthony Bau, Yonatan Belinkov, Hassan
  Sajjad, Nadir Durrani, and James Glass. 2019.
\newblock \href {https://www.aaai.org/ojs/index.php/AAAI/article/view/5063}
  {Neurox: A toolkit for analyzing individual neurons in neural networks}.
\newblock In \emph{AAAI Conference on Artificial Intelligence (AAAI)}.

\bibitem[{de~Marneffe et~al.(2021)de~Marneffe, Manning, Nivre, and
  Zeman}]{de-marneffe-etal-2021-universal}
Marie-Catherine de~Marneffe, Christopher~D. Manning, Joakim Nivre, and Daniel
  Zeman. 2021.
\newblock \href {https://doi.org/10.1162/coli_a_00402} {{U}niversal
  {D}ependencies}.
\newblock \emph{Computational Linguistics}, 47(2):255--308.

\bibitem[{Devlin et~al.(2018)Devlin, Chang, Lee, and
  Toutanova}]{https://doi.org/10.48550/arxiv.1810.04805}
Jacob Devlin, Ming-Wei Chang, Kenton Lee, and Kristina Toutanova. 2018.
\newblock \href {https://doi.org/10.48550/ARXIV.1810.04805} {Bert: Pre-training
  of deep bidirectional transformers for language understanding}.

\bibitem[{Fayyaz et~al.(2021)Fayyaz, Aghazadeh, Modarressi, Mohebbi, and
  Pilehvar}]{fayyaz-etal-2021-models}
Mohsen Fayyaz, Ehsan Aghazadeh, Ali Modarressi, Hosein Mohebbi, and
  Mohammad~Taher Pilehvar. 2021.
\newblock \href {https://doi.org/10.18653/v1/2021.blackboxnlp-1.29} {Not all
  models localize linguistic knowledge in the same place: A layer-wise probing
  on {BERT}oids{'} representations}.
\newblock In \emph{Proceedings of the Fourth BlackboxNLP Workshop on Analyzing
  and Interpreting Neural Networks for NLP}, pages 375--388, Punta Cana,
  Dominican Republic. Association for Computational Linguistics.

\bibitem[{Haspelmath(2001)}]{haspelmath2001sae}
Martin Haspelmath. 2001.
\newblock The european linguistic area: Standard average european.
\newblock In Wulf~Oesterreicher Martin~Haspelmath and Wolfgang Raible, editors,
  \emph{Language Typology and Language Universals, Handbücher zur Sprach- und
  Kommunikationswissenschaft}, pages 1492--1510. Mouton de Gruyter, Berlin.

\bibitem[{Hewitt and Liang(2019)}]{Hewitt2019DesigningAI}
John Hewitt and Percy Liang. 2019.
\newblock Designing and interpreting probes with control tasks.
\newblock \emph{ArXiv}, abs/1909.03368.

\bibitem[{Hoover et~al.(2020)Hoover, Strobelt, and
  Gehrmann}]{Hoover2020exBERTAV}
Benjamin Hoover, Hendrik Strobelt, and Sebastian Gehrmann. 2020.
\newblock exbert: A visual analysis tool to explore learned representations in
  transformer models.
\newblock In \emph{ACL}.

\bibitem[{Jawahar et~al.(2019)Jawahar, Sagot, and
  Seddah}]{jawahar-etal-2019-bert}
Ganesh Jawahar, Beno{\^\i}t Sagot, and Djam{\'e} Seddah. 2019.
\newblock \href {https://doi.org/10.18653/v1/P19-1356} {What does {BERT} learn
  about the structure of language?}
\newblock In \emph{Proceedings of the 57th Annual Meeting of the Association
  for Computational Linguistics}, pages 3651--3657, Florence, Italy.
  Association for Computational Linguistics.

\bibitem[{Koto et~al.(2021)Koto, Lau, and Baldwin}]{Koto2021DiscoursePO}
Fajri Koto, Jey~Han Lau, and Tim Baldwin. 2021.
\newblock Discourse probing of pretrained language models.
\newblock In \emph{NAACL}.

\bibitem[{Kurfal{\i} and {\"O}stling(2021)}]{kurfali-ostling-2021-probing}
Murathan Kurfal{\i} and Robert {\"O}stling. 2021.
\newblock \href {https://doi.org/10.18653/v1/2021.repl4nlp-1.2} {Probing
  multilingual language models for discourse}.
\newblock In \emph{Proceedings of the 6th Workshop on Representation Learning
  for NLP (RepL4NLP-2021)}, pages 8--19, Online. Association for Computational
  Linguistics.

\bibitem[{Loshchilov and Hutter(2017)}]{DBLP:journals/corr/abs-1711-05101}
Ilya Loshchilov and Frank Hutter. 2017.
\newblock \href {http://arxiv.org/abs/1711.05101} {Fixing weight decay
  regularization in adam}.
\newblock \emph{CoRR}, abs/1711.05101.

\bibitem[{Madsen et~al.(2021)Madsen, Reddy, and Chandar}]{Madsen2021PosthocIF}
Andreas Madsen, Siva Reddy, and A.~P.~Sarath Chandar. 2021.
\newblock Post-hoc interpretability for neural nlp: A survey.
\newblock \emph{ArXiv}, abs/2108.04840.

\bibitem[{Merlo(2019)}]{merlo-2019-probing}
Paola Merlo. 2019.
\newblock \href {https://doi.org/10.18653/v1/W19-4817} {Probing word and
  sentence embeddings for long-distance dependencies effects in {F}rench and
  {E}nglish}.
\newblock In \emph{Proceedings of the 2019 ACL Workshop BlackboxNLP: Analyzing
  and Interpreting Neural Networks for NLP}, pages 158--172, Florence, Italy.
  Association for Computational Linguistics.

\bibitem[{Mikhailov et~al.(2021)Mikhailov, Taktasheva, Sigdel, and
  Artemova}]{mikhailov-etal-2021-rusenteval}
Vladislav Mikhailov, Ekaterina Taktasheva, Elina Sigdel, and Ekaterina
  Artemova. 2021.
\newblock \href {https://aclanthology.org/2021.bsnlp-1.6} {{R}u{S}ent{E}val:
  Linguistic source, encoder force!}
\newblock In \emph{Proceedings of the 8th Workshop on Balto-Slavic Natural
  Language Processing}, pages 43--65, Kiyv, Ukraine. Association for
  Computational Linguistics.

\bibitem[{Nichols(2007)}]{nichols-2007}
Johanna Nichols. 2007.
\newblock What, if anything, is typology?

\bibitem[{Pires et~al.(2019)Pires, Schlinger, and
  Garrette}]{pires2019multilingual}
Telmo Pires, Eva Schlinger, and Dan Garrette. 2019.
\newblock How multilingual is multilingual bert?
\newblock \emph{arXiv preprint arXiv:1906.01502}.

\bibitem[{Ravishankar et~al.(2019{\natexlab{a}})Ravishankar, {\O}vrelid, and
  Velldal}]{Ravishankar2019ProbingMS}
Vinit Ravishankar, Lilja {\O}vrelid, and Erik Velldal. 2019{\natexlab{a}}.
\newblock Probing multilingual sentence representations with x-probe.
\newblock In \emph{RepL4NLP@ACL}.

\bibitem[{Ravishankar et~al.(2019{\natexlab{b}})Ravishankar, {\O}vrelid, and
  Velldal}]{ravishankar-etal-2019-probing}
Vinit Ravishankar, Lilja {\O}vrelid, and Erik Velldal. 2019{\natexlab{b}}.
\newblock \href {https://doi.org/10.18653/v1/W19-4318} {Probing multilingual
  sentence representations with {X}-probe}.
\newblock In \emph{Proceedings of the 4th Workshop on Representation Learning
  for NLP (RepL4NLP-2019)}, pages 156--168, Florence, Italy. Association for
  Computational Linguistics.

\bibitem[{Rogers et~al.(2020)Rogers, Kovaleva, and
  Rumshisky}]{rogers-etal-2020-primer}
Anna Rogers, Olga Kovaleva, and Anna Rumshisky. 2020.
\newblock \href {https://doi.org/10.1162/tacl_a_00349} {A primer in
  {BERT}ology: What we know about how {BERT} works}.
\newblock \emph{Transactions of the Association for Computational Linguistics},
  8:842--866.

\bibitem[{Sahin et~al.(2020)Sahin, Vania, Kuznetsov, and
  Gurevych}]{Sahin2020LINSPECTORMP}
G{\"o}zde~G{\"u}l Sahin, Clara Vania, Ilia Kuznetsov, and Iryna Gurevych. 2020.
\newblock Linspector: Multilingual probing tasks for word representations.
\newblock \emph{Computational Linguistics}, 46:335--385.

\bibitem[{Tikhonova et~al.(2022)Tikhonova, Mikhailov, Pisarevskaya, Malykh, and
  Shavrina}]{tikhonova_mikhailov_pisarevskaya_malykh_shavrina_2022}
Maria Tikhonova, Vladislav Mikhailov, Dina Pisarevskaya, Valentin Malykh, and
  Tatiana Shavrina. 2022.
\newblock \href {https://doi.org/10.1017/S1351324922000225} {Ad astra or
  astray: Exploring linguistic knowledge of multilingual bert through nli
  task}.
\newblock \emph{Natural Language Engineering}, page 1–30.

\bibitem[{Voloshina et~al.(2022)Voloshina, Serikov, and
  Shavrina}]{voloshinaetal}
Ekaterina Voloshina, Oleg Serikov, and Tatiana Shavrina. 2022.
\newblock Is neural language acquisition similar to natural?a chronological
  probing study.

\bibitem[{Wallace et~al.(2019)Wallace, Tuyls, Wang, Subramanian, Gardner, and
  Singh}]{Wallace2019AllenNLPIA}
Eric Wallace, Jens Tuyls, Junlin Wang, Sanjay Subramanian, Matt Gardner, and
  Sameer Singh. 2019.
\newblock Allennlp interpret: A framework for explaining predictions of nlp
  models.
\newblock \emph{ArXiv}, abs/1909.09251.

\bibitem[{Wu and Dredze(2019)}]{wu2019beto}
Shijie Wu and Mark Dredze. 2019.
\newblock Beto, bentz, becas: The surprising cross-lingual effectiveness of
  bert.
\newblock \emph{arXiv preprint arXiv:1904.09077}.

\bibitem[{Şahin et~al.(2019)Şahin, Vania, Kuznetsov, and
  Gurevych}]{https://doi.org/10.48550/arxiv.1903.09442}
Gözde~Gül Şahin, Clara Vania, Ilia Kuznetsov, and Iryna Gurevych. 2019.
\newblock \href {https://doi.org/10.48550/ARXIV.1903.09442} {Linspector:
  Multilingual probing tasks for word representations}.

\end{thebibliography}
\bibliographystyle{acl_natbib}
\clearpage
\appendix
\onecolumn
\section{Appendix}
\label{sec:appendix}

\subsection{Languages information and statistic}
\label{sec:languages}

After the processing of Universal Dependencies, 104 languages left. Here is a table with the languages, their families, and the number of examples that we used in the experiments.

\begin{table}[htp!]
\fontsize{11}{11.5}\selectfont
\resizebox{\linewidth}{!}{
\begin{tabular}{@{}lll|lll@{}}
\toprule
\textbf{Language}          & \textbf{Family}          & \textbf{Examples} & \textbf{Language}            & \textbf{Family}         & \textbf{Examples} \\ \midrule
Afrikaans         & Indo-European   & 19646    & Komi Zyrian         & Uralic         & 5682     \\
Akkadian          & Afro-Asiatic    & 15037    & Korean              & Koreanic       & 3314     \\
Akuntsu           & Tupian          & 79       & Kurmanji            & Indo-European  & 9134     \\
Albanian          & Indo-European   & 380      & Latin               & Indo-European  & 1048162  \\
Amharic           & Afro-Asiatic    & 7166     & Latvian             & Indo-European  & 393694   \\
Ancient Greek     & Indo-European   & 566076   & Ligurian            & Indo-European  & 2304     \\
Apurina           & Arawakan        & 176      & Lithuanian          & Indo-European  & 81462    \\
Arabic            & Afro-Asiatic    & 484604   & Livvi               & Uralic         & 835      \\
Armenian          & Indo-European   & 37117    & Low Saxon           & Indo-European  & 623      \\
Assyrian          & Afro-Asiatic    & 152      & Manx                & Indo-European  & 13536    \\
Bambara           & Mande           & 4829     & Marathi             & Indo-European  & 6340     \\
Basque            & -               & 191646   & Mbya Guarani        & Tupian         & 5503     \\
Beja              & Afro-Asiatic    & 347      & Moksha              & Uralic         & 3133     \\
Belarusian        & Indo-European   & 445666   & Munduruku           & Tupian         & 164      \\
Bengali           & Indo-European   & 156      & Naija               & Atlantic-Congo & 42928    \\
Bhojpuri          & Indo-European   & 1525     & North Sami          & Uralic         & 21639    \\
Breton            & Indo-European   & 5206     & Norwegian           & Indo-European  & 631651   \\
Bulgarian         & Indo-European   & 238822   & Old Church Slavonic & Indo-European  & 118508   \\
Buryat            & Mongolic-Khitan & 6832     & Old East Slavic     & Indo-European  & 153393   \\
Catalan           & Indo-European   & 305522   & Old French          & Indo-European  & 45115    \\
Chinese           & Sino-Tibetan    & 18865    & Persian             & Indo-European  & 183678   \\
Classical Chinese & Sino-Tibetan    & 93864    & Polish              & Indo-European  & 860418   \\
Coptic            & Afro-Asiatic    & 22150    & Portuguese          & Indo-European  & 197481   \\
Croatian          & Indo-European   & 193156   & Romanian            & Indo-European  & 543203   \\
Czech             & Indo-European   & 831540   & Russian             & Indo-European  & 270189   \\
Danish            & Indo-European   & 104906   & Sanskrit            & Indo-European  & 25885    \\
Dutch             & Indo-European   & 241808   & Scottish Gaelic     & Indo-European  & 27907    \\
English           & Indo-European   & 414215   & Serbian             & Indo-European  & 94856    \\
Erzya             & Uralic          & 17458    & Skolt Sami          & Uralic         & 989      \\
Estonian          & Uralic          & 557773   & Slovak              & Indo-European  & 218032   \\
Faroese           & Indo-European   & 21133    & Slovenian           & Indo-European  & 286196   \\
Finnish           & Uralic          & 624845   & Spanish             & Indo-European  & 660046   \\
French            & Indo-European   & 686410   & Swedish             & Indo-European  & 213496   \\
Galician          & Indo-European   & 15878    & Tagalog             & Austronesian   & 380      \\
German            & Indo-European   & 311259   & Tamil               & Dravidian      & 12602    \\
Gothic            & Indo-European   & 99064    & Tatar               & Turkic         & 250      \\
Greek             & Indo-European   & 49364    & Thai                & Tai-Kadai      & 612      \\
Guajajara         & Tupian          & 409      & Tupinamba           & Tupian         & 593      \\
Hebrew            & Afro-Asiatic    & 112866   & Turkish             & Turkic         & 746291   \\
Hindi             & Indo-European   & 321197   & Turkish German      & Indo-European  & 21160    \\
Hungarian         & Uralic          & 41102    & Ukrainian           & Indo-European  & 139882   \\
Icelandic         & Indo-European   & 503790   & Upper Sorbian       & Indo-European  & 5241     \\
Indonesian        & Austronesian    & 47426    & Urdu                & Indo-European  & 96902    \\
Irish             & Indo-European   & 93264    & Uyghur              & Turkic         & 40174    \\
Italian           & Indo-European   & 395470   & Warlpiri            & Pama-Nyungan   & 128      \\
Javanese          & Austronesian    & 290      & Welsh               & Indo-European  & 17026    \\
Kaapor            & Tupian          & 99       & Western Armenian    & Indo-European  & 71303    \\
Karelian          & Uralic          & 1475     & Wolof               & Atlantic-Congo & 21518    \\
Karo              & Tupian          & 1845     & Xibe                & Tungusic       & 4226     \\
Kazakh            & Turkic          & 15082    & Yakut               & Turkic         & 213      \\
Kiche             & Indo-European   & 11534    & Yoruba              & Atlantic-Congo & 1151     \\
Komi Permyak      & Uralic          & 325      & Yupik               & Eskimo-Aleut   & 2281
\end{tabular}
}
\end{table}

\onecolumn
\subsection{mBERT's layers $F_1$ scores for all languages and categories.}
\label{sec:heatmap_mbert}
\begin{figure*}[!htbp]
    \centering
    \includegraphics[width=\linewidth]{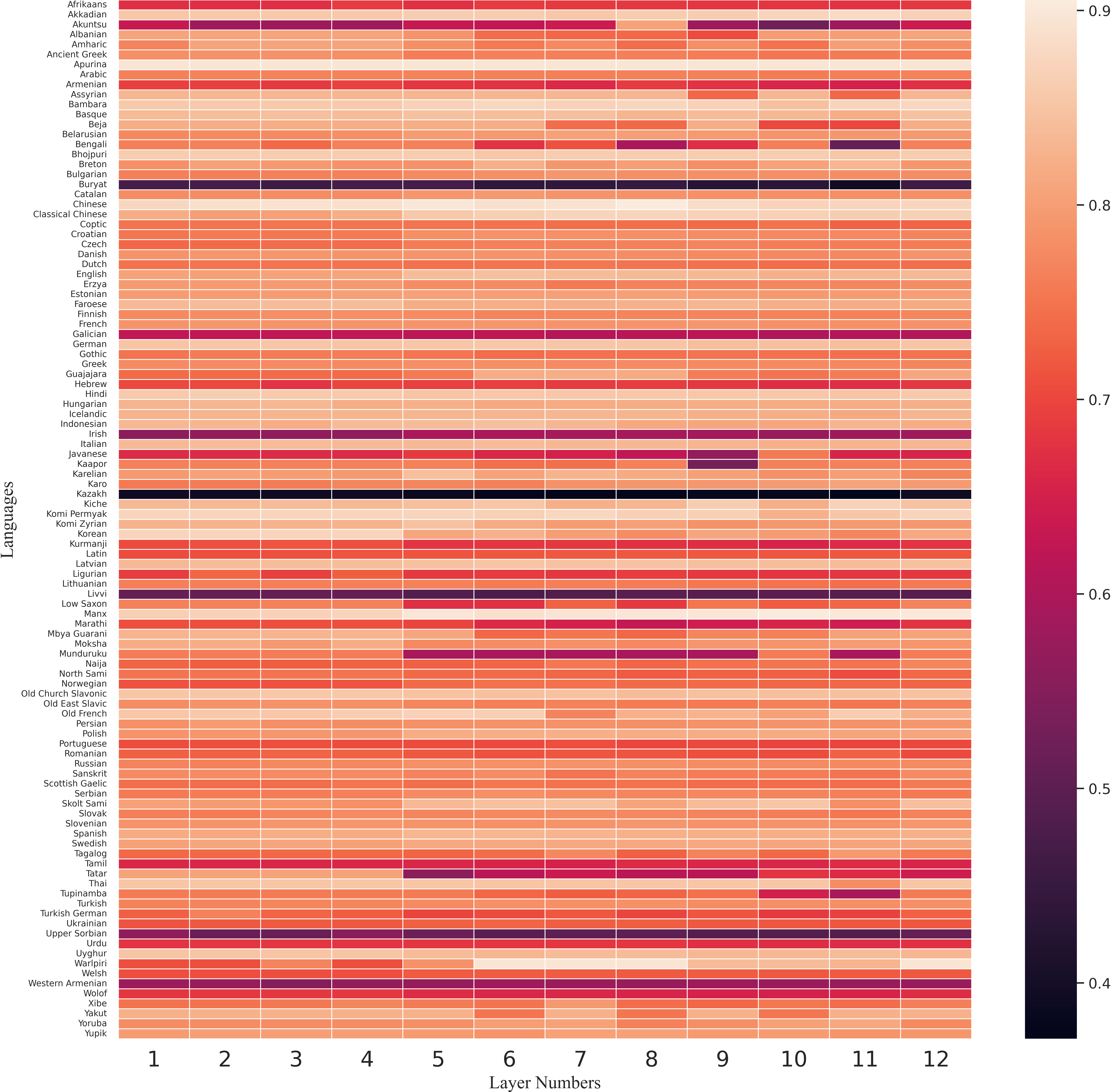}
    \caption{Distribution of average model scores by layers for languages measured on all categories.}
    \label{fig:vitalya_layers}
\end{figure*}

\onecolumn
\subsection{XLM-R results grouped by languages and average feature probing score on all layers}
\label{sec:heatmap_xlm_r}
\begin{figure*}[!htb]
    \centering
    \includegraphics[width=\textwidth]{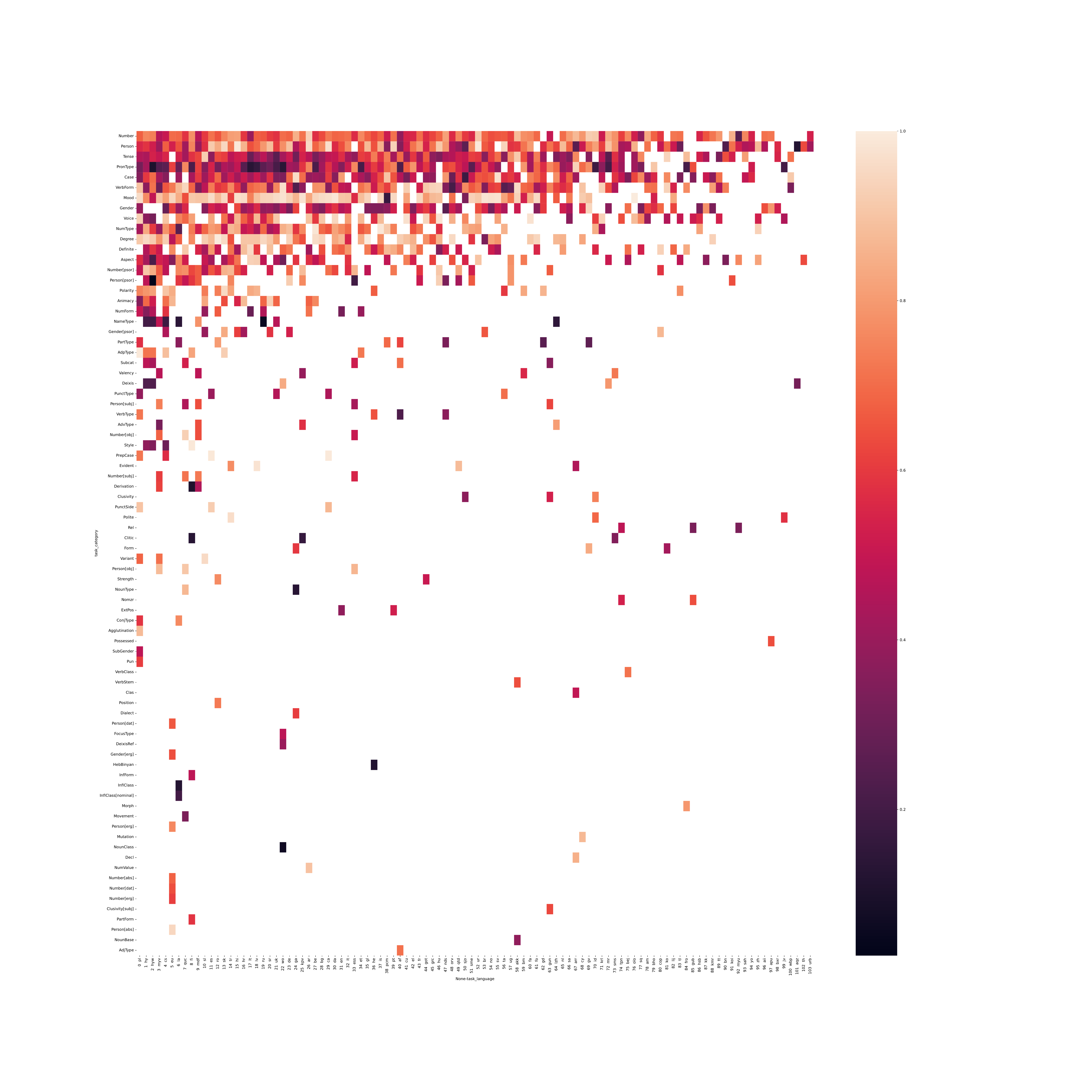}
    \caption{XLM-R results grouped by languages and average feature probing score on all layers}
    \label{fig:xlmr_hmap}
\end{figure*}

\onecolumn
\subsection{Model's layers $F_1$ scores for all languages  on category "Number".}
\label{sec:heatmap_number}
\begin{figure*}[!htbp]
    \centering
    \includegraphics[width=\textwidth]{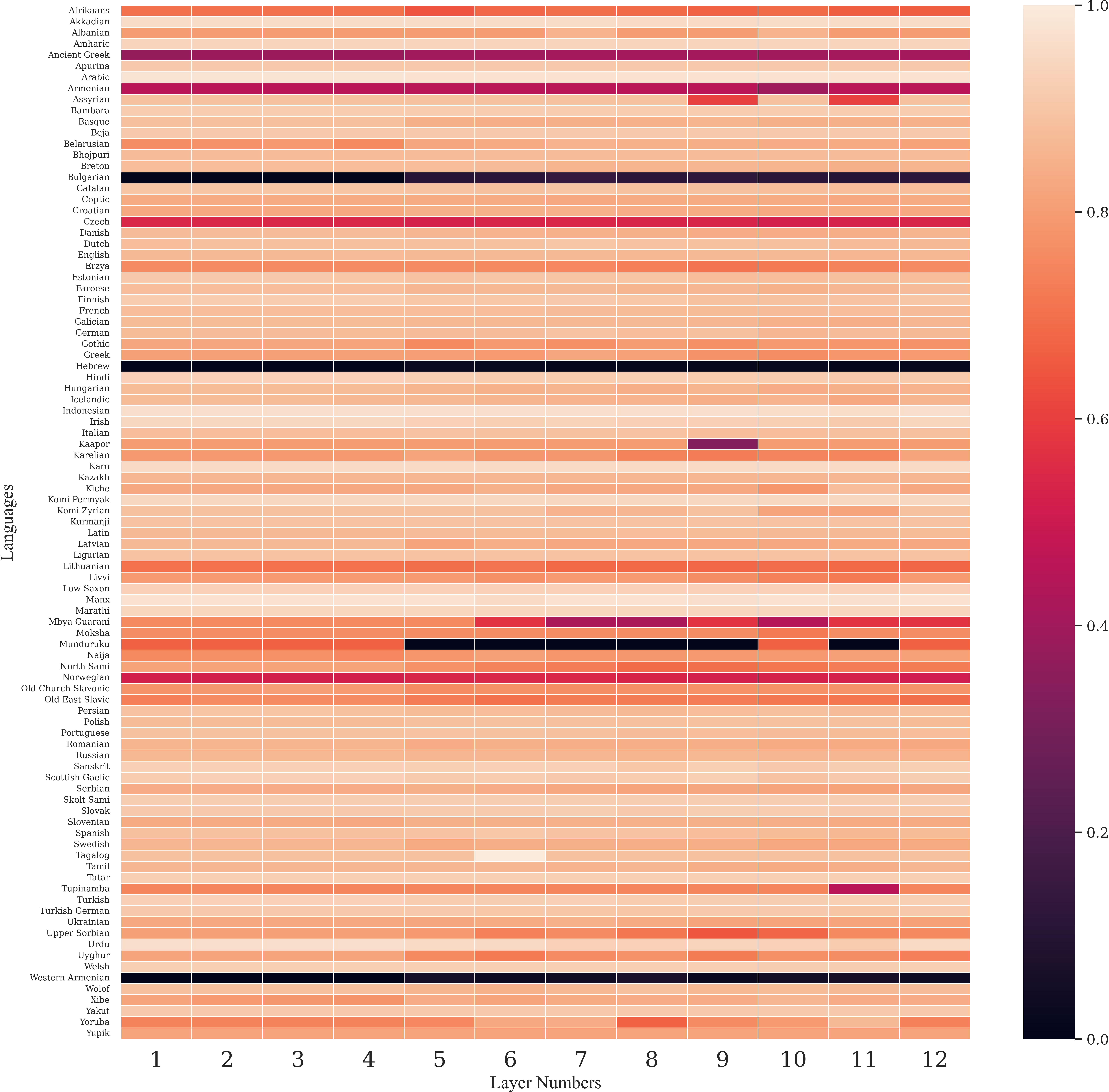}
    \caption{Distribution of model scores by layers for languages measured on category \textit{Number}.}
\end{figure*}

\onecolumn
\subsection{Model's layers $F_1$ scores for all languages  on category "PronType".}
\label{sec:heatmap_prontype}
\begin{figure*}[!htbp]
    \centering
    \includegraphics[width=\textwidth]{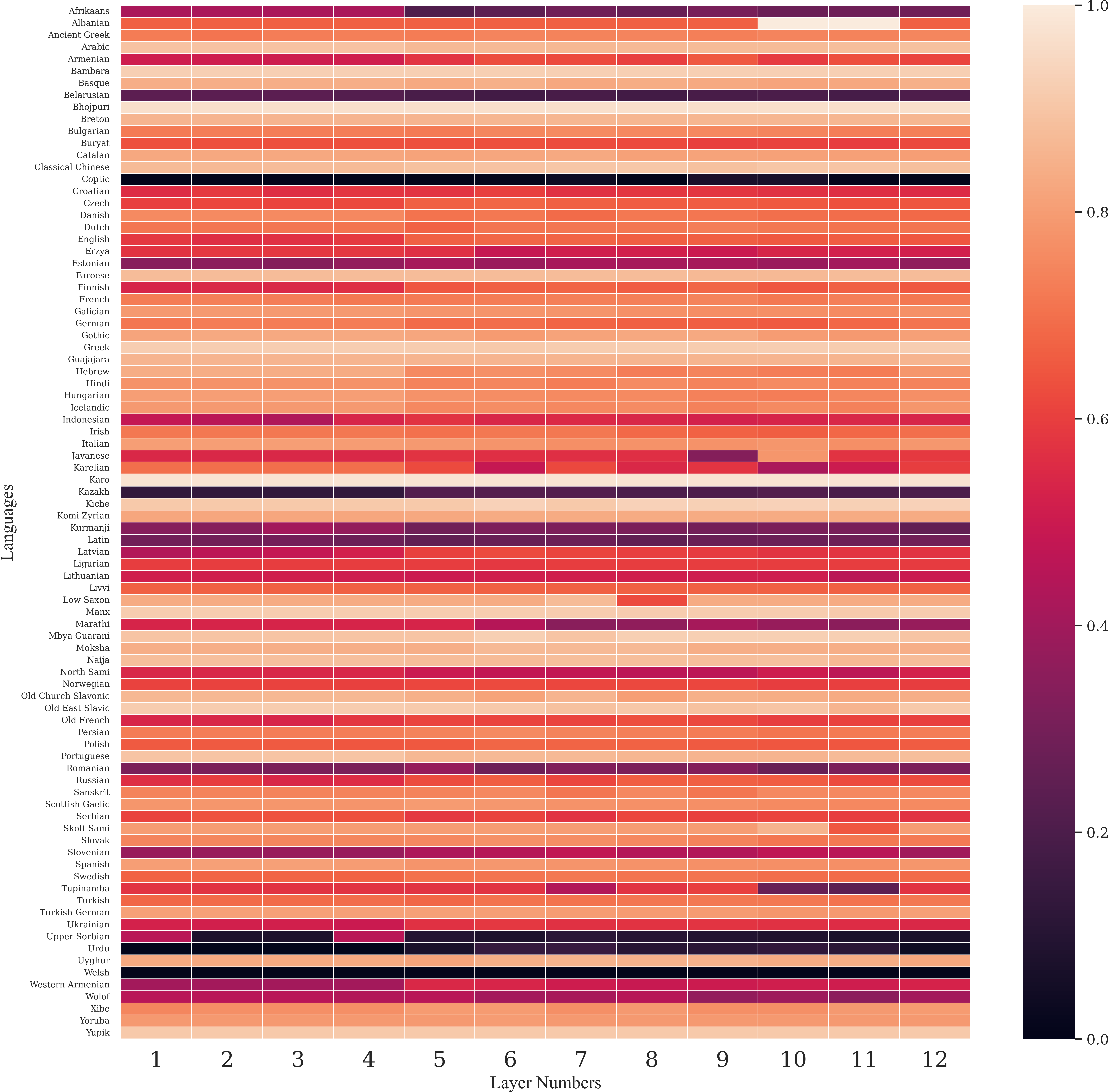}
    \caption{Distribution of model scores by layers for languages measured on category \textit{PronType}.}
\end{figure*}

\end{document}